\documentclass[10pt,twocolumn,letterpaper]{article}
\usepackage[breaklinks=true,bookmarks=false]{hyperref}
\usepackage{wacv}
\usepackage{times}
\usepackage{epsfig}
\usepackage{graphicx}

\usepackage{amsmath}
\usepackage{amssymb}

\newcommand\Tstrut{\rule{0pt}{2.3ex}}   
\newcommand\Bstrut{\rule[-1.3ex]{0pt}{0pt}}   

\usepackage{subfig}
\usepackage{multicol}
\usepackage{multirow}
\usepackage{diagbox}
\usepackage{graphicx}
\usepackage{comment}

\newcommand\blfootnote[1]{%
  \begingroup
  \renewcommand\thefootnote{}\footnote{#1}%
  \addtocounter{footnote}{-1}%
  \endgroup
}

\def\wacvPaperID{0142} 

\wacvfinalcopy 

\ifwacvfinal
\def\assignedStartPage{1} 
\fi

\usepackage[english]{babel}
\addto\extrasenglish{
    
}
\addto\extrasenglish{
    
}

\ifwacvfinal
\else
\fi

\begin{document}

\title{Quantitative Analysis of Image Classification Techniques for Memory-Constrained Devices}

\renewcommand*{\thefootnote}{\fnsymbol{footnote}}

\author{Sebastian M{\"u}ksch$^\dagger$$^*$\\
{\tt\small sebastian.mueksch@gmail.com}
\and
Theo Olausson$^\dagger$$^*$\\
{\tt\small tolausso@exseed.ed.ac.uk}
\and
John Wilhelm$^\dagger$$^*$\\
{\tt\small john.r.wilhelm@outlook.com}
\and
Pavlos Andreadis$^\dagger$\\
{\tt\small pavlos.andreadis@ed.ac.uk}
}
\maketitle

\begin{abstract} 

Convolutional Neural Networks, or CNNs, are the state of the art for image classification, but typically come at the cost of a large memory footprint. This limits their usefulness in applications relying on embedded devices, where memory is often a scarce resource. Recently, there has been significant progress in the field of image classification on such memory-constrained devices, with novel contributions like the ProtoNN, Bonsai and FastGRNN algorithms. These have been shown to reach up to 98.2\% accuracy on optical character recognition using MNIST-10, with a memory footprint as little as 6KB. However, their potential on more complex multi-class and multi-channel image classification has yet to be determined. In this paper, we compare CNNs with ProtoNN, Bonsai and FastGRNN when applied to 3-channel image classification using CIFAR-10. For our analysis, we use the existing Direct Convolution algorithm to implement the CNNs memory-optimally and propose new methods of adjusting the FastGRNN model to work with multi-channel images. We extend the evaluation of each algorithm to a memory size budget of 8KB, 16KB, 32KB, 64KB and 128KB to show quantitatively that Direct Convolution CNNs perform best for all chosen budgets, with a top performance of 65.7\% accuracy at a memory footprint of 58.23KB.

\end{abstract}

\section{Introduction}
\label{sec:introduction}
A\blfootnote{$^*$ Co-first authors}pplying image classification in the real world is a task which comes with several innate challenges:
occlusion, intra-class variability, varying lighting conditions and, more recently, adversarial examples form only the start of a long list of problems which need to be overcome. Significant progress has been made towards solving this open problem via deep learning, in particular in the form of Convolutional Neural Networks, or CNNs \cite{NIPS2012_4824}. However, prominent models to have achieved state-of-the-art performance tend to be very large (e.g. \cite{simonyan2014deep}).
This limits the feasibility of applying CNNs to carry out image classification on memory-constrained devices. One is left with little option but to offload the inference to a data centre, which brings several challenges of its own. For example, immediate technological challenges with offloading include reliability, its affect on the overall system cost \cite{offloading-matters} and a potential worsening of battery life and latency \cite{norman_2019}. The rise of 5G internet connectivity could mitigate some of these concerns by, for instance, offering greater reliability and stability of  communication. However, several interesting areas of application may simply be incompatible with offloading. A matter of privacy \cite{6322974}, such as in surveillance systems, or latency and battery life, such as in implantable medical devices to detect, for instance, seizures \cite{seizure-detection}, may give reason to avoid offloading.

To circumvent the issues associated with offloading inference onto a data centre, one may instead carry out inference directly on the embedded device by using models with a small enough memory footprint.
While constructing powerful models with memory size in mind is no easy task, the desire to avoid offloading has sparked diverse streams of research targeting equally diverse applications.
As far as image classification goes, experimental results have thus far been centred around the MNIST data set \cite{mnist}. This data set consists of $28\times28$ single-channel images containing handwritten digits (0-9). As each image only takes up $28 \times 28 \times 1 = 784$ bytes, with 1 byte per pixel, it leaves a majority of the memory available to the model even when memory is highly constrained. However, in the present day the usefulness of this data set is limited, due to two main reasons:
\begin{itemize}
\itemsep-1em
     \item Recent works are starting to reach saturating levels of performance, i.e. above $99\%$ test set accuracy of models fitting within 2KB of memory \cite{direct-convolutions}. \\
     \item The single-channel nature of the
data set may bias the state of the art towards methods which may not generalise well when
the input images consist of several channels, e.g. coloured images in RGB or HSV encoding.
\end{itemize}

Motivated by these insights, this paper compares the state-of-the-art methods for memory-constrained image classification on a significantly more challenging data set for the domain: CIFAR-10 \cite{cifar-10}. To present a thorough evaluation, only memory constraints are considered and the matters of latency and energy efficiency are left for future work.

\autoref{sec:background} introduces each analysed model and highlights its perceived strengths and weaknesses in simpler applications. The CIFAR-10 data set is discussed briefly in \autoref{sec:data-set}. Next, \autoref{sec:methodology} details how each method can be applied to image classification, describing also how to search for the best model for a given memory size budget. Notably, this section suggests new methods of adjusting the FastGRNN model to work optimally with multi-channel images.  \autoref{sec:exps} then details and compares the experimental results for each method when they are allowed to use up to 8KB, 16KB, 32KB, 64KB or 128KB of memory. Finally, \autoref{sec:conclusions} summarises the findings of this paper and outlines future work.

\section{Literature review}
\label{sec:background}

In this section, we introduce the recent models from the literature which we set out to analyse in this paper. All of these are chosen as they represent the state-of-the-art for their respective approaches of memory-constrained classification.

We will discuss \emph{Direct Convolutions} as the state-of-the-art CNN approach to memory-constrained image classification \cite{direct-convolutions}, \emph{ProtoNN} as a method to reduce memory requirements of a k-NN model \cite{protonn}, \emph{Bonsai} as a decision-tree based model \cite{bonsai} and finally \emph{FastGRNN}, a state-of-the-art tiny RNN model \cite{fastgrnn}.

\subsection{Direct Convolutions}
\label{sec:direct-colvolutions}

The \emph{Direct Convolution} neural network proposed in \cite{direct-convolutions}
is a method which significantly reduces the memory overhead of using Convolutional Neural Networks
through clever re-use of memory. Memory used to store the pixels of an input feature map
is progressively replaced with the activations of the layer as the inputs become stale, i.e. all activations depending
on the input pixel have been computed \cite{direct-convolutions}.

Though deceptively simple,
this method is made significantly more complicated when a layer increases the channel depth: that is,
when the number of channels in the input is strictly less than that in the output. In such scenarios,
naively processing the pixels in row-major
order would cause the memory to be freed in a fragmented manner, making it difficult to store the output activations \cite{direct-convolutions}. To deal with this case, a \emph{herringbone} strategy is proposed where the pixels are traversed in alternating row- and column-major
order,
which provably uses the minimal amount of space \cite{direct-convolutions} in addition to that taken up by the input image.

As of writing, this method holds the current record performance
($99.15\%$ test accuracy) on 10-class MNIST classification for models with memory footprints on the order of 
kilobytes, and
it does so with a model of only 2KB of memory \cite{direct-convolutions}. However, no results
for a data set other than MNIST are presented in \cite{direct-convolutions}, nor were any such results identified elsewhere in the literature.

\subsection{ProtoNN}
\label{sec:protonn-lit-rev}

A different approach to object recognition on resource-scare systems is given in \cite{protonn}.
This paper introduces the \emph{ProtoNN} algorithm, which takes inspiration from the familiar \emph{k-Nearest-Neighbours} 
method (described amongst other places in \cite{kNN}). ProtoNN performs inference analogously to k-NN, i.e. assigning a data point to a class based on the most frequent class of its nearest neighbours. It distinguishes itself, however, by requiring several orders of magnitude less space and time. This is achieved by learning a small set of
informative \textit{prototype} data points to compare against at time of inference, along with a sparse projection onto
low-dimensional space \cite{protonn}.

Compared to Direct Convolutions \cite{direct-convolutions}, ProtoNN achieves a less impressive $95.88\%$ accuracy on 10-class MNIST classification and requires 64KB of memory to do so \cite{protonn}. However, unlike in \cite{direct-convolutions}, ProtoNN is compared against
a wider range of data sets in \cite{protonn}, including a 2-class version of CIFAR on which they achieve $76.35\%$ accuracy at 16KB.
As such, there is more evidence that it will obtain comparable performance in a wider range of tasks.

\subsection{Bonsai}
\label{sec:bonsai-lit-review}

A decision-tree based algorithm for resource-constrained machine learning dubbed \emph{Bonsai} is proposed in \cite{bonsai}, which along with learning a non-linear tree also learns a low-dimensional projection matrix. The model size is kept small by training a single tree rather than an entire forest on the 
low-dimensional projected data and by making sure that the learned projection matrix is sparse \cite{bonsai}.

Though the authors carry out experiments targeting several data sets, the results are mainly compared to
those of pruned version of large networks, rather than architectures which directly target the resource-constrained systems.
Though it may have been state-of-the-art at the time of its publication, Bonsai's $97.01\%$ 10-class MNIST test set accuracy at 84KB \cite{bonsai} is now outperformed by Direct Convolutions \cite{direct-convolutions}.

Like with ProtoNN in \cite{protonn}, a more
varied set of experiments than just 10-class MNIST classification is carried out with Bonsai in \cite{bonsai}. One of the obtained results is $73.02\%$ accuracy on the 2-class version 
of CIFAR by only using up 2KB of memory; at 16KB it reaches $76.64\%$, just about beating out ProtoNN \cite{protonn}. As such,
we believe Bonsai to still be an interesting contender in the memory-constrained image classification space.

\subsection{FastGRNN}
\label{sec:bg-fastgrnn}

Gating mechanisms have a rich history and have long been used to stabilize training and improve performance of RNNs, such as for example in Long Short-Term Memory units, or LSTMs, \cite{lstm}. However, gates add extra parameters to the model and thus increase its memory footprint, and as a result the gating mechanisms used in LSTMs have since been refined by GRUs \cite{GRU} and UGRNNs \cite{UGRNN}. \emph{FastGRNN} is a gated recurrent neural network proposed in \cite{fastgrnn} which continues along this line of producing smaller gating mechanisms.

FastGRNN uses only two prediction parameters to compute both the gate activation and the hidden state, and the model can then further be compressed through quantization or keeping the two prediction parameters low-rank, sparse or both.
To allow for this compression, FastGRNN uses a three-stage training process split evenly over the training epochs. In the first stage, the sparsity constraints are ignored while a low-rank version of the parameters are learned. In the second stage, the sparsity structure is learned along with the support for the prediction parameters. In the third and final stage, the support set of the parameters is frozen and the model is then fine-tuned \cite{fastgrnn}.

As it is a recurrent architecture, FastGRNN's primary use case
is handling sequential data, such as speech. Indeed, one of the most impressive results of the original paper is that
the model is able to accurately capture the wakeword ``Hey Cortana" with a model size of only 1KB \cite{fastgrnn}. 
However, somewhat surprisingly, FastGRNN is also apt at image classification, achieving $98.2\%$ accuracy with
a 6KB model on a pixel-by-pixel version of 10-class MNIST \cite{fastgrnn}. As such, it makes for an interesting contender in our analysis.

\section{Data set and Task} 
\label{sec:data-set}


The models which we will be examining in this analysis, described in \autoref{sec:background}, have been tested on multiple data sets and different number of target classes. For instance, Direct Convolution has been tested on MNIST 10-class classification \cite{direct-convolutions} and ProtoNN on MINST/CIFAR 2-class classification \cite{protonn}. In either of these cases the dimensionality of the feature space is either small (MNIST has $28\times28$ single-channel images) or the number of target classes is low. For our analysis, we therefore choose the CIFAR-10 data set \cite{cifar-10} as we argue that it is significantly more complex and thus allows us to establish a new benchmark for the models described in \autoref{sec:background}.

The CIFAR-10 data set consists of 60,000 $32\times32$ 3-channel colour images, divided into 10 classes such as airplane, automobile and dog \cite{cifar-10}. 
It comes split into 50,000 images for training and 10,000 for testing. The training images are split into five training batches of 10,000 images each and the test images are kept in one test batch of 10,000 images \cite{cifar-10}.
The test batch is balanced with respect to the 10 classes, featuring 1,000 randomly-selected images from each class, but the five training batches may be unbalanced \cite{cifar-10}. However, together the training batches contain exactly 5,000 images from each class \cite{cifar-10}.

For our analysis, we join the training batches into one set and then extract a hold-out validation set consisting of exactly 1000 images of each class. Thus, we are left with 40,000 images for training. This validation set is used to set, for example, hyper-parameters and to select the best models for each algorithm described in \autoref{sec:background}. This ensures that the final test set accuracy reported for each model is an unbiased estimate of its generalisation accuracy.

\section{Methodology}
\label{sec:methodology}

This section will detail how each of the methods outlined in \autoref{sec:background} can be applied to solve the 3-channel image classification problem considered in this paper. For each method, when applicable, possible model structures, parameterisations and sequencing of input data will be outlined. Additionally, practical algorithms for searching for the optimal configuration of the method for a given memory budget will be described. 

\subsection{Direct Convolutions}
\label{sec:meth-direct-conv}
\begin{table*}[h]
  \centering
\begin{small}
  \begin{sc}
\begin{tabular}{|l|l||l|l|}
 \hline
 \multicolumn{2}{|l||}{\textbf{Layer Name \& Abbreviation}}  & \textbf{Fixed Parameters} & \textbf{Variable Parameters} \Tstrut\Bstrut\\
 \hline
  \hline
Average Pooling 2D&$A$& $pool\_size=(2,2)$ & - \Tstrut\Bstrut\\
\hline
Maximum Pooling 2D&$M$& $pool\_size=(2,2)$ & - \Tstrut\Bstrut\\
\hline
Dense \emph{with} activation&$D$& $activation=ReLU,$  &$output\_dim\in\{16, 32, 64\}$ \Tstrut\Bstrut\\
\hline
Dense \emph{without} activation&$D^*$& $activation=None,\ output\_dim=10$  & - \Tstrut\\
\hline
2D Convolution&$C_1$& $activation=None,\ strides = (1,1),$ & $output\_dim\in\{4, 6, 8, 10, 12, 16, 32, 64\}$ \Tstrut\\
 &&$padding=valid$ & $kernel\_size=(k,k),\ k \in\{1,3,5\}$\\
\hline
Depthwise 2D Convolution&$C_2$ &$activation=ReLU,\ strides = (1,1),$ & $output\_dim\in\{4, 6, 8, 10, 12, 16, 32, 64\}$ \Tstrut\\
 &&$ padding = valid,\ multiplier=1$ & $kernel\_size=(k,k),\ k \in\{3,5\}$\\
\hline
Dropout&$Dr$&$rate=0.1$ & - \Tstrut\Bstrut\\
 \hline
\end{tabular}
\caption{Direct Convolution candidate layers with abbreviations and their respective parameter spaces. \cite{direct-convolutions}}
\label{table:conv-layer-explanation}
\end{sc}
\end{small}
\end{table*}


\begin{table*}[h]
  \centering
\begin{footnotesize}
\begin{sc}
\begin{tabular}{|l|l|l|l|l|l|}
\hline
\multicolumn{6}{|c|}{\textbf{Serial Convolutional Neural Netowrk Architectures}} \Tstrut\Bstrut\\
\hline
\hline
$A, C, C, C, M, Dr, D^{*}$& $A, C, M, D, Dr, D^{*}$ & $A, C, D, Dr, D^{*}$ & $A, C, M, C, Dr, D^{*}$& $A, C, C, M, Dr, D^{*}$ & $A, C, C, Dr, D^{*}$  \Tstrut\Bstrut\\
\hline
$A, D, D, D, Dr, D^{*}$ & $A, C, M, C, D, Dr, D^{*}$ &$A, C, D, D, Dr, D^{*}$ & $A, C, M, D, D, D^{*}$ & $A, C, C, M, D, D^{*}$ & $A, C, C, D, D^{*}$ \Tstrut\Bstrut\\
\hline
$A, C, M, C, C, Dr, D^{*}$ & $A, C, C, M, C, Dr, D^{*} $ & $A, D, D, Dr, D^{*}$  &$A, C, C, C, Dr,  D^{*}$&\multicolumn{1}{c}{~}  &\multicolumn{1}{c}{~}  \Tstrut\Bstrut\\
\cline{1-4}
\end{tabular}

\caption{Enumeration of serial model architectures considered in Direct Convolution approach. See \autoref{table:conv-layer-explanation} for abbreviations. Note that $C$ can be either $C_1$ or $C_2$. \cite{direct-convolutions}}
\label{table:direct-conv-layer-comb}
\end{sc}
\end{footnotesize}
\end{table*}

As outlined in \autoref{sec:direct-colvolutions}, Direct Convolutions form a protocol implementing CNNs memory-optimally \cite{direct-convolutions}. The main result in \cite{direct-convolutions}, which introduced this method, was a 99.15\% classification accuracy on the single-channel images of the 10-class MNIST data set. We will seek to extend these results to the 3-channel images of the CIFAR-10 data set \cite{cifar-10}.

In order to obtain 99.15\% classification accuracy on 10-class MNIST, a sampling-based neural architecture search was performed \cite{direct-convolutions}. The candidate layers for this search, their fixed parameters and the variable parameters are given in \autoref{table:conv-layer-explanation}. The 16 possible combinations of these layers searched over in \cite{direct-convolutions} are given in \autoref{table:direct-conv-layer-comb}. Given the strong performance of these models, we will also use them in our analysis.

As in \cite{direct-convolutions}, searching for an optimal configuration is done by firstly generating all of the possible models, i.e. all combinations of architectures and variable parameters, then calculating the memory requirements for each. From the list of generated models, a given number of models satisfying the memory size budget, are sampled. Given empirical and theoretical evidence that randomly searching for hyper-parameters is more efficient than a guided or grid search \cite{randomsearch}, we believe this sampling approach to be reasonable. Each of the selected models is partially trained for 5 epochs and the best model after this limited training is identified.  Finally, the identified model is trained fully using early stopping. The model resulting from this process is considered the optimal one.

\subsection{ProtoNN}
\label{sec:prot-meto}
As introduced in \autoref{sec:protonn-lit-rev}, ProtoNN is a classification algorithm which learns a sparse, low-dimensionality projection and a set of prototype data points which can be considered as the `training data' in the normal inference procedure associated with the $k-$NN algorithm \cite{kNN}. Given that the algorithm does not assume any spacial relationship between elements of the feature vectors, single- as well as multi-channel images can be flattened out into feature vectors. The two parameters of this algorithm are the dimensionality of the projected space as well as the number of prototypes to learn, which both have an effect on the model size. Further, a sparsity constraint on the learned matrices can be imposed in order to reduce the complexity of the model. An additional parameter of this method is $\gamma$ from the Gaussian kernel acting as a similarity function in the method.

To obtain the best performing model, a trade-off between the dimensionality of the projected space, the number of prototypes to learn and the sparsity constraint is required. A grid search over these parameters along with $\gamma$ and the learning rate of the training is performed over values which respect the memory bound. The scale of this parameter search is tuned by considering the step size between parameters in the grid search.

\subsection{Bonsai}
\label{sec:bonsai-methodology}

As described in \autoref{sec:bonsai-lit-review}, the Bonsai method is a decision-tree-based algorithm that learns a low-dimensional projection matrix alongside the decision tree itself. Since the Bonsai model takes an entire image as a singular vector, with all channels concatenated \cite{bonsai}, Bonsai is already built for multi-channel image classification.

As a Bonsai model is parameterised by the depth of the decision tree and the dimensionality of the projection matrix and both are integer-valued \cite{bonsai}, we exhaustively search over these two parameters. We fix the tree depth and train models with varying dimensionality of the projection matrix. We start with a dimensionality of 1 and increment the dimensionality with each new model. We stop when we reach a model that does no longer fit into our largest memory size budget of 128KB. After that, we increment the tree depth and repeat the process.

Initially, we run the aforementioned search with the tree depth ranging from 1 to 6. As we find that there are models with higher tree depth which still fit within 128KB, we continue the search for a tree depth of 7 and 8. We stop at a tree depth of 8 because we find the model performance to deteriorate despite increasing model sizes, all described in \autoref{sec:bonsai-experiments}.



\subsection{FastGRNN}
\label{sec:fastgrnn-methodology}

As introduced in \autoref{sec:bg-fastgrnn}, FastGRNN is a recurrent neural network architecture which has
shown surprising potential in simple image classification domains \cite{fastgrnn}.
However, to our knowledge we are the first to apply it to a domain with multi-channel images. This raises the
question of how to model the input data as a time series in order to benefit from the recurrent nature of the network.

In simple single-channel images such as those found in MNIST \cite{mnist},
the input data can be turned into a time series
either by feeding the network with a single pixel at a time (as done in \cite{fastgrnn}) or by feeding the network with a group of pixels, e.g. a full row or column, at a time (as done in examples provided by \cite{edgemlsrc}).
In either case,
multi-channel images complicate this process by introducing an implicit trade-off between proximity in the time series between pixels which lie close together within a single channel (e.g. adjacent pixels in the red channel) and pixels which lie close together across the channels (e.g. the first pixels in each of the three channels).

\begin{figure}[htpb]
  \centering
  \begin{tabular}{@{}c@{}}
     \includegraphics[width=0.45\textwidth]{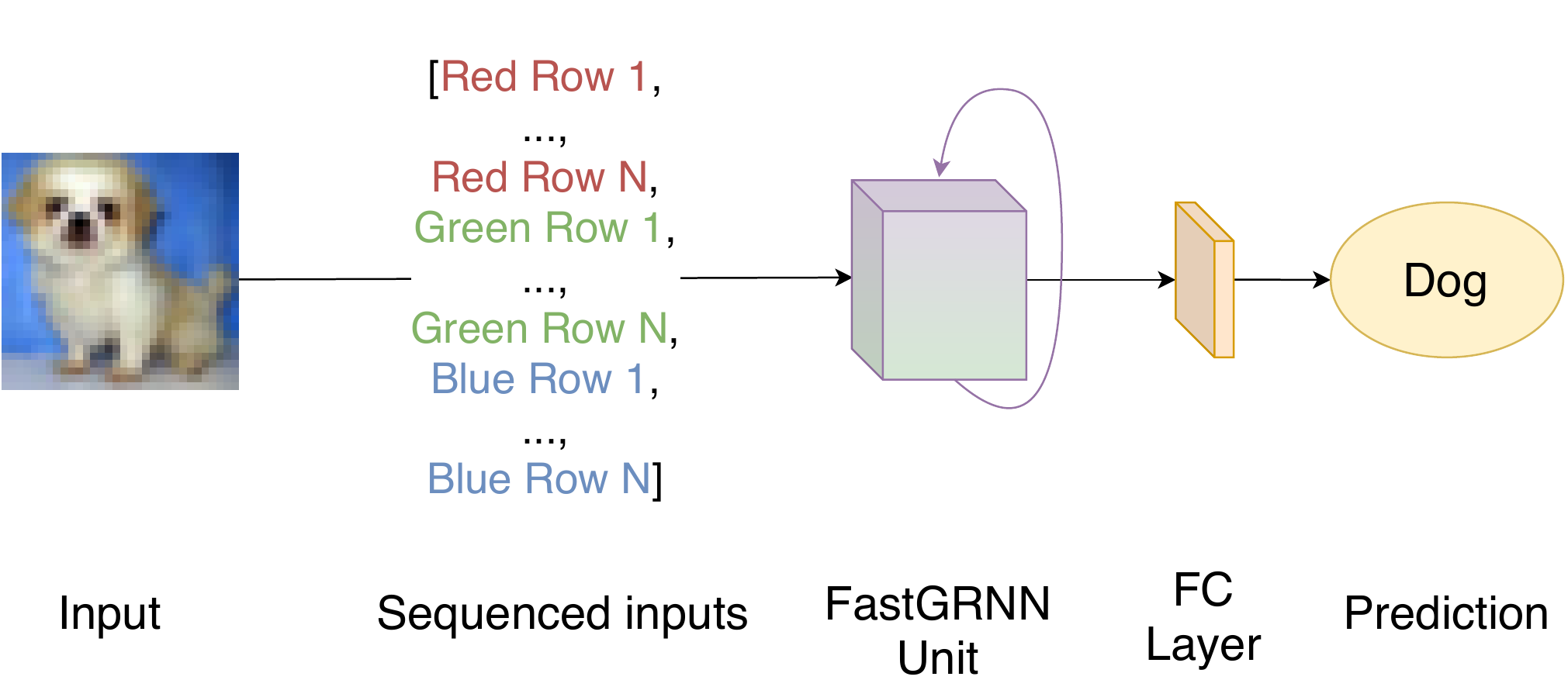} \\[\abovecaptionskip]
    \small (a) Row-major FastGRNN data feed.
  \end{tabular}

  \begin{tabular}{@{}c@{}}
     \includegraphics[width=0.45\textwidth]{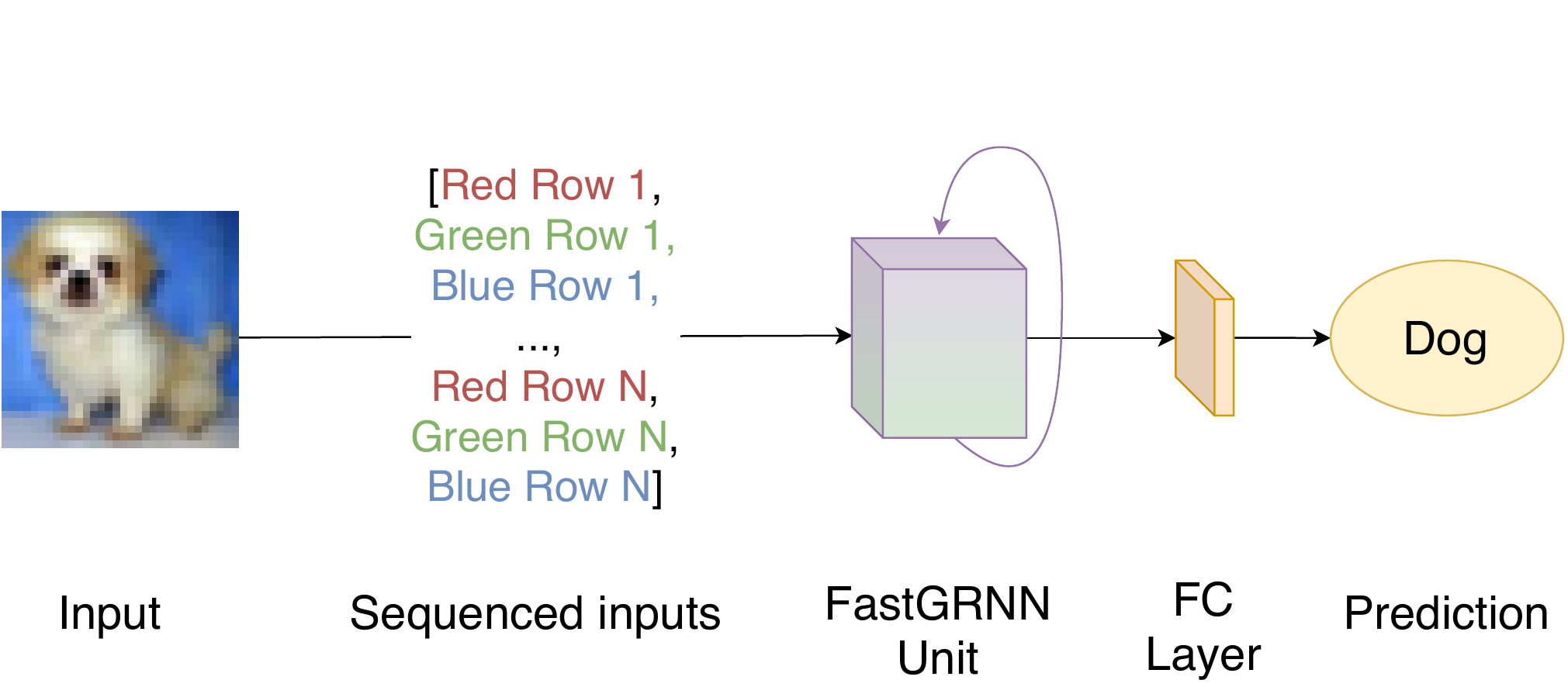} \\[\abovecaptionskip]
    \small (b) Channel-major FastGRNN data feed.
  \end{tabular}

  \begin{tabular}{@{}c@{}}
     \includegraphics[width=0.45\textwidth]{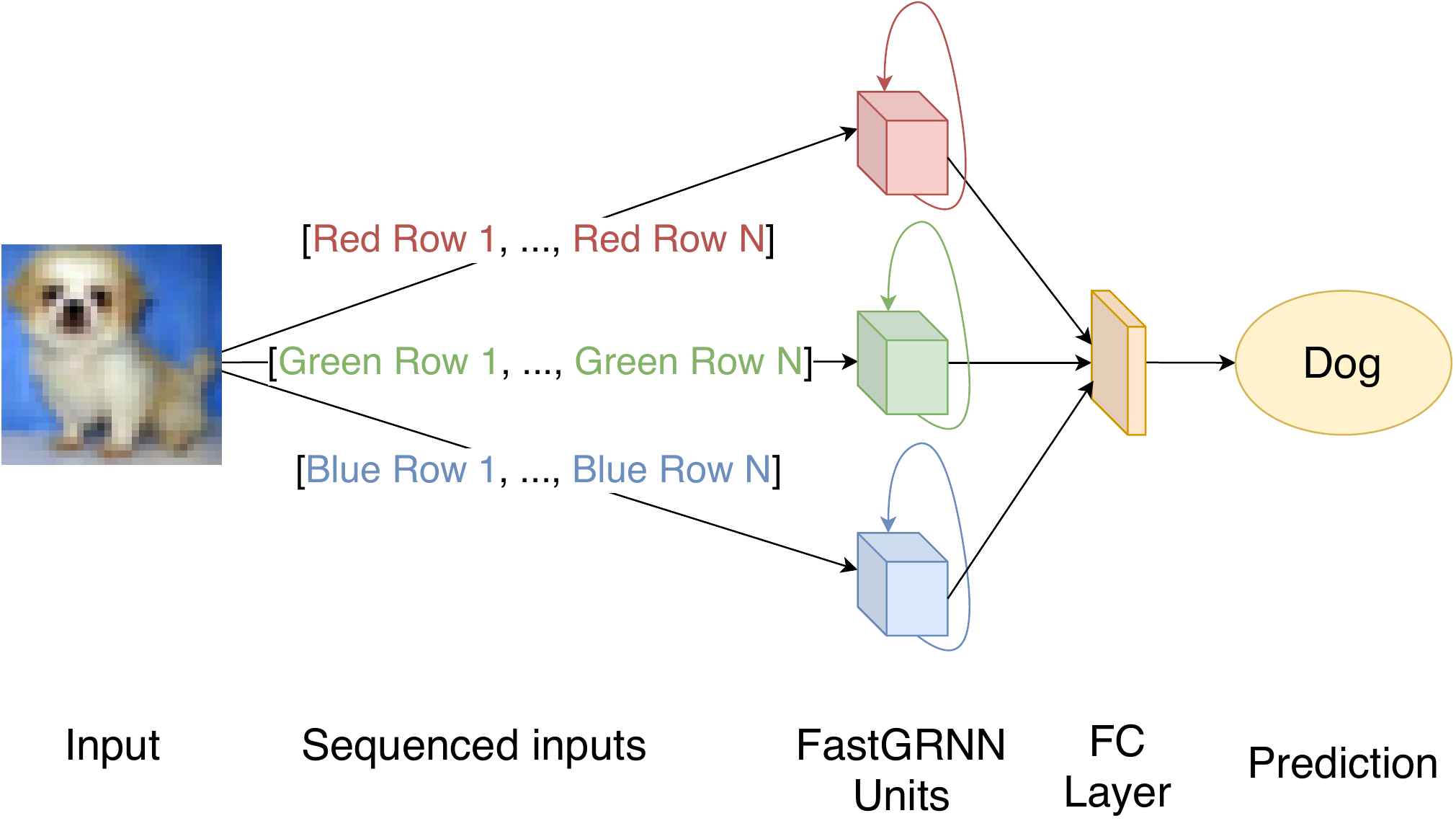}\\[\abovecaptionskip]
    \small (c) Proposed Multi-FastGRNN architecture.
  \end{tabular}

  \caption{The three methods for applying FastGRNN to multi-channel images. The input image is a sample from CIFAR-10 \cite{cifar-10}.}\label{fig:architectures}
\end{figure}

In this paper, we devote significant attention to comparing FastGRNN's performance on CIFAR-10 image classification for different
modes of sequencing the input data. We propose three different methods for classifying multi-channel images
with the FastGRNN architecture. These share the basic assumption that each data point in the time series is one row of one channel in the input, but then differ in how they feed these data points into the network:

\begin{itemize}
    \setlength{\parskip}{0em}
    
    \item \textbf{Row-major}: Feed the data into a single FastGRNN unit, followed by a fully-connected layer, starting first with all red rows, then all green rows and finally all blue rows. See \autoref{fig:architectures} (a).
    
    \item \textbf{Channel-major}: Feed the data into a single FastGRNN unit, followed by a fully-connected layer, starting with the first red row, the first green row, the first blue row, then the second red row, the second green row and the second blue row, etc. until the last red row, the last green row, the last blue row. See \autoref{fig:architectures} (b).
    
    \item \textbf{Multi-FastGRNN}: Feed the data into three separate FastGRNN units, one for each channel, followed by a fully connected layer. Feed each unit with the rows of the channel corresponding to the unit, in order. See \autoref{fig:architectures} (c).
    
\end{itemize}

For an RNN, learning features from several elements in its input sequence is tied to their \emph{temporal latency}, i.e. distance between them in the sequence. With the row-major method, we focus on features that relate the pixels in the first red row to those in the second red row and so on, where it takes time to see the next channel. With the channel-major method, we focus on features that relate pixels in the first rows of each channel, where it takes time to see the next row. As such there is a trade-off between setting up for intra- and inter-channel features.
With the Multi-FastGRNN architecture we seek to circumvent this trade-off
by explicitly separating the channels and training one FastGRNN unit per channel.
To retain the ability to still learn cross-channel features, we then concatenate the outputs of each unit and feed the combined output into a fully-connected layer.

The size of a FastGRNN unit is predominantly determined by its input dimensionality and its hidden dimensionality. As briefly mentioned in \autoref{sec:background}, the model can then be further compressed if the predictor parameters $U$ and $W$ \cite{fastgrnn} are kept low-rank and sparse, or if the parameters are quantized.
In this analysis, we focus on the hidden dimensionality and the sparsity of $U$ and $W$ to tune the size of the models, to accommodate for the large amount of size budgets and input sequencing methods to investigate. We thus keep the prediction parameters full-rank and do not carry out quantization, a trade-off which we do not believe to be a cause for concern, as our analysis entails comparatively large memory budgets for the domain (up to 128KB).

\label{sec:experiments}







\section{Experiments}
\label{sec:exps}
In this section we set up and perform experiments
following the methodology outlined in \autoref{sec:methodology}.
For the ProtoNN, Bonsai, and the FastGRNN methods we use
the versions included in the 
the EdgeML library \cite{edgemlsrc}, while for Direct Convolutions we base our experiments on the software provided by \cite{cnnsrc}.
To ensure that the reported performance of each method is as accurate
of a reflection of its potential in the test domain,
we devote significant time to individually optimising each method
in our experiments.


\subsection{Direct Convolutions}
 As detailed in \autoref{sec:meth-direct-conv}, a sampling based neural architecture search is performed in order to obtain the best performaning models for this method. In particular, we sample 750 models for each memory budget and partially train the candidate models for 5 epochs. To train each candidate model we used the Adam optimiser \cite{adam} with an adaptive learning rate, which was initialized at 0.005 and decayed by 0.95 each epoch. Full training in this experiment is training the model for 100 epochs with early stopping with a patience of 5.  The selected models after partial training are given in \autoref{table:conv-best-models} for each budget along with the size the model and the test accuracy of the final model after full training. Note that for full training, models were trained with the same optimiser as used for the partial training.
 
 \begin{table}[htpb]
\vskip 3mm
\begin{center}
\begin{small}
\begin{sc}
\resizebox{\columnwidth}{!}{
\begin{tabular}{|c||cc|}
 \hline
 \textbf{Budget} & \textbf{Model Architecture}& \textbf{Test Accuracy} \Tstrut\Bstrut\\
   & \textbf{and parameterisation}& \textbf{[Size]} \Tstrut\Bstrut\\
 \hline
 \hline
 \multirow{2}{*}{$\leq$ 8KB}  & $A, C_2(16, (3,3)), C_1(8, (3,3)), $ & 0.604\Tstrut \\
    &  $\hdots, C_1(32,(3,3)), M, Dr, D^{*}$ &[5.39KB]  \Bstrut\\
 \hline
 \multirow{2}{*}{$\leq$ 16KB} & $A, C_1(6, (3,3)), C_1(32, (1,1)), M,$ & 0.629 \Tstrut\\
      &  $\hdots,  C_2(64,(3,3)), Dr, D^{*}$ &  [8.65KB] \Bstrut\\
 \hline
 \multirow{2}{*}{$\leq$ 32KB} & $A, C_1(8, (1,1)), C_2(16, (3,3)),$ & 0.643  \Tstrut\\
   &  $\hdots, C_1(64, (5,5)), M, Dr, D^{*}$ & [19.91KB] \Bstrut\\
 \hline
 $\leq 64$KB & $A, C_1(64, (3,3)), M, C_1(64, (1,1)),$ & 0.657 \Tstrut\\
 \ $\leq 128$KB  &  $\hdots,  C_1(64,(5,5)), Dr, D^{*}$ & [58.23KB] \Bstrut\\
 \hline
\end{tabular}
}
\end{sc}
\end{small}
\caption{The best network architectures for the Direct Convolution method. See \autoref{table:conv-layer-explanation} for abbreviations. Convolutional layers $C_1$ and $C_2$ are followed by the value of their variable arguments, in the order $output\_dim$ then $kernel\_size$.}
\label{table:conv-best-models}
\end{center}
\vskip -3mm
\end{table}
 
 The results of this experiment show that the Direct Convolutions method performs modestly at the classification task of interest. Moreover, the best model for each memory budget is not necessarily close to using all of the available memory size.

%
%


\subsection{ProtoNN}
Recall that \autoref{sec:prot-meto} outlines a grid search for obtaining the best performing ProtoNN model for a given memory budget. For this search method, the dimensionality of the projected space and the number of prototypes both vary in the values 2, 4, 8, 16, 32 and 64 in order to obtain exponentially larger models. For initial experimentation, the sparsity value was fixed at 1.0. For the parameter $\gamma$ \cite{protonn}, we use $1.5 \times 10^n$, where we range $n$ from -4 to 4 in integer steps. We vary the parameter $\gamma$ this extensively due the method's sensitivity to it \cite{protonn}. Finally, the learning rate ranged in 0.1, 0.01 and 0.001. We train the models for 100 epochs with early stopping with a patience of 10.

The models in this experiment ranged in size from 24.67KB to 805.38KB. The best model for the appropriate memory budgets is given in \autoref{table:best-proton}. Note further that of all of the models trained in this experiment, the best model with respect to validation accuracy was of size 791.16KB, with a final test accuracy of 0.147. 

From this experiment it can be concluded that the ProtoNN method is able to learn somewhat how to solve this multi-class problem, but only to a limited extent.

\begin{table}[h]
\vskip 3mm
\begin{center}
\begin{small}
\begin{sc}
\resizebox{\columnwidth}{!}{
\begin{tabular}{|c||c|}
\hline
\Tstrut\Bstrut
\textbf{Budget} & $\leq$ 32KB, $\leq$ 64KB and $\leq$ 128KB   \\

\hline
\hline
\Tstrut\Bstrut
\textbf{Sparsity,\ Learning Rate} &  1.0, 0.01  \\
\textbf{Projection Dim.,\ No. Prototypes} &  2, 4  \\

\Tstrut
\textbf{Test Accuracy} &0.147 \\
\textbf{[Size]} &\ \ \  [24.77KB]\ \ \  \\

\hline
\end{tabular}
}
\end{sc}
\end{small}
\caption{Best ProtoNN model for each memory budget with its parameterisation.}
\label{table:best-proton}
\end{center}
\vskip -3mm
\end{table}

\subsection{Bonsai}
\label{sec:bonsai-experiments}


We search over possible Bonsai models as described in \autoref{sec:bonsai-methodology}, by fixing the depth of the Bonsai tree and varying the dimensionality of the projection matrix.

For all Bonsai experiments we choose a learning rate of 0.01, using the Adam optimiser \cite{adam}. and a batch size of 224, the square root of the original 50,000 training samples \cite{cifar-10}. For the Bonsai-specific hyperparameters, we set the sigmoid sharpness to 1 and set a regulariser of $10^{-3}$ for predictor parameters W and V and also branching parameter $\theta$. For W and V we set a sparsity of 0.3 and for $\theta$ a sparsity of 0.62. Finally, we set a regulariser of $10^{-4}$ with sparsity 0.2 for projection parameter Z, all based on \cite{bonsai}.

We train all models for 200 epochs. We do not perform early stopping due to Bonsai's sequence of distinct training phases \cite{bonsai}, but for each memory size budget choose the model with the highest validation accuracy after the 200 training epochs.

\begin{figure}
    \centering
    \includegraphics[width=\columnwidth]{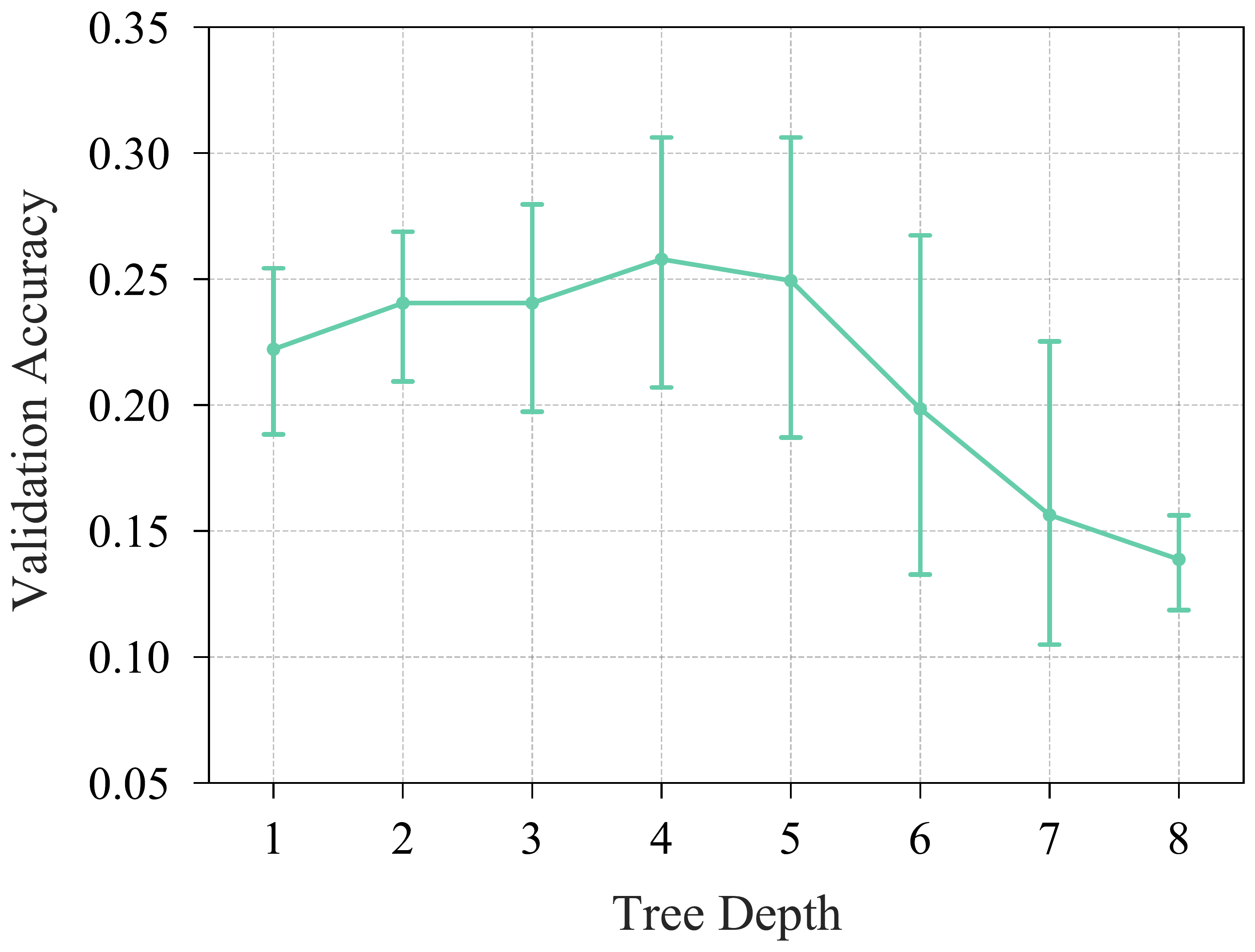}
    \caption{Average validation accuracy of 129 Bonsai models for given tree depth. All models require less than or equal to 128KB of memory.}
    \label{fig:bonsai-avg-val-accuracy}
\end{figure}

After searching over tree depths of 1 to 6 as described in \autoref{sec:bonsai-methodology}, we find that deeper Bonsai decision trees can still fit within the maximum 128KB memory size budget. We search over models with tree depth 7 and 8 according to the search method described in \autoref{sec:bonsai-methodology}. However, as \autoref{fig:bonsai-avg-val-accuracy} demonstrates, the validation accuracies of trees with depth 7 and 8 are, on average, significantly lower than for shallower Bonsai trees. We attribute this to a necessarily smaller dimensionality of the projection matrix required to fit deeper trees into the memory budgets. For example, a Bonsai tree of depth 5 can utilise a projection matrix with dimensionality up to 16 and achieving up to 37.6\% validation accuracy with a memory size of 94.52KB. Compare this to a tree of depth 8 which can utilise a projection matrix up to only dimensionality 4 and only achieving a maximum of 16.3\% while requiring 89.25KB of memory. We therefore stop exploring deeper Bonsai trees and summarise the best models in \autoref{table:bonsai-best-models}.

\begin{table}[h]
\vskip 3mm
\begin{center}
\begin{small}
\begin{sc}
\resizebox{\columnwidth}{!}{
\begin{tabular}{|c||ccccc|}
\hline
\Tstrut\Bstrut
\textbf{Budget} & $\leq$ 8KB & $\leq$ 16KB & $\leq$ 32KB & $\leq$ 64KB & $\leq$ 128KB \\
\hline
\hline
\Tstrut\Bstrut
\textbf{Depth} & 5 & 2 & 2 & 3 & 5 \\
\textbf{Dim.} & 1 & 3 & 6 & 11 & 12 \\
\Bstrut
\textbf{Test Accuracy} & $0.149$ & $0.153$ & $0.221$ & $0.325$ & $0.377$ \\
\Bstrut
\textbf{[Size]} & [7.88KB] & [15.43KB] & [30.85KB] & [60.86KB] & [94.52KB] \\
\hline
\end{tabular}
}
\end{sc}
\end{small}
\caption{Best Bonsai models for each memory budget with given tree depth and dimensionality of projection matrix.}
\label{table:bonsai-best-models}
\end{center}
\vskip -3mm
\end{table}



\subsection{FastGRNN}
\label{sec:experiments-fastgrnn}

We begin our FastGRNN experiments by constructing several candidate models for each size budget.
For the row- and channel-major models we vary the density of $U$ and $W$ independently between 0.1, 0.2 and 0.3 based on recommendations by the EdgeML library \cite{edgemlsrc}. However, we also consider fully dense models where the densities of $U$ and $W$ are both 1.0. We then construct models by varying the hidden dimensionality in steps of 15 between 0 and 225.
Finally, to make training tractable, we discard all but three models per size budget, keeping the models which approach the size budget most closely.


For Multi-FastGRNN we fix the density of $U$ to 1.0 as we find the compression of $U$ to have a negligible effect on the model size due to the small hidden dimensionality per unit, which we vary in steps of 5 between 0 and 100. We again discard all but three candidate models per budget. However, as the search does not yield three models for the 8KB size budget, we manually add a dense model with hidden dimensionality 12 and a sparse model with hidden dimensionality 14 and density 0.1 for $W$.


We are left with 45 models to train: three per size budget and data feed method. Due to the large number of models to compare, we fix the update and gate non-linearities to the hyperbolic tangent and sigmoid function, respectively. These are identified as good defaults by \cite{fastgrnn}.

\begin{table}[h]
\vskip 3mm
\begin{small}
\begin{center}
\begin{sc}
\resizebox{\columnwidth}{!}{
\begin{tabular}{c|c||ccccc|}
\cline{2-7}
\Tstrut\Bstrut
&\textbf{Budget} & $\leq$ 8KB & $\leq$ 16KB & $\leq$ 32KB & $\leq$ 64KB & $\leq$ 128KB \\
\cline{2-7}
\noalign{\vskip\doublerulesep
         \vskip-\arrayrulewidth}
\noalign{\vskip\doublerulesep
         \vskip-\arrayrulewidth}
         
\hline
\Tstrut\Bstrut
\multirow{6}{*}{{\rotatebox[origin=c]{90}{\textbf{Row}}}}&\textbf{Input Dim.} & 32 & 32 & 32 & 32 & 32 \\
&\textbf{Hidden Dim.} & 45 & 75 & 120 & 150 & 210 \\
\Tstrut
&\textbf{Density(W)} & 0.2 & 0.1 & 0.1 & 0.1 & 0.1 \\
\Tstrut
&\textbf{Density(U)} & 0.2 & 0.2 & 0.2 & 0.3 & 0.3 \\
\Tstrut
&\textbf{Test Accuracy} & $0.471$ & $0.515$ & $0.541$ & $0.572$ & $0.587$ \\
\Bstrut
&\textbf{[Size]} & [7.57KB] & [14.23KB] & [31.17KB] & [63.56KB] & [118.50KB] \\
\hline
\hline
\Tstrut\Bstrut
\multirow{6}{*}{{\rotatebox[origin=c]{90}{\textbf{Channel}}}}&\textbf{Input Dim.} & 32 & 32 & 32 & 32 & 32 \\
&\textbf{Hidden Dim.} & 45 & 60 & 105 & 150 & 150 \\
\Tstrut
&\textbf{Density(W)} & 0.2 & 0.3 & 0.3 & 0.1 & 0.1 \\
\Tstrut
&\textbf{Density(U)} & 0.2 & 0.3 & 0.2 & 0.3 & 0.3 \\
\Tstrut
&\textbf{Test Accuracy} & $0.482$ & $0.533$ & $0.553$ & $0.589$ & $0.589$  \\
\Bstrut
&\textbf{[Size]} & [7.57KB] & [15.80KB] & [30.07KB] & [63.56KB] & [63.56KB]  \\
\hline
\hline
\Tstrut\Bstrut
\multirow{4}{*}{{\rotatebox[origin=c]{90}{\textbf{Multi}}}}&\textbf{Input Dim.} & 32 & 32 & 32 & 32 & 32 \\
&\textbf{Hidden Dim.} & 12 & 20 & 35 & 55 & 90 \\
\Tstrut
&\textbf{Density(W)} & 1.0 & 1.0 & 0.3 & 1.0 & 0.3 \\
\Tstrut
&\textbf{Density(U)} & 1.0 & 1.0 & 1.0 & 1.0 & 1.0 \\
\Tstrut
&\textbf{Test Accuracy} & $0.447$ & $0.477$ & $0.527$ & $0.558$ & $0.558$ \\
\Bstrut
&\textbf{[Size]} & [7.94KB] & [15.06KB] & [28.75KB] & [63.87KB] & [124.09KB] \\
\hline

\end{tabular}
}
\end{sc}
\caption{Best FastGRNN models for each memory budget and data feed method. The hidden and input dimensionalities for the Multi-FastGRNN are per FastGRNN unit, replicated for each channel. Note that for Multi-FastGRNN the 64KB and 128KB models obtained the same test accuracy when rounded to 3 decimal places, but the latter is still included as it obtained a higher validation accuracy.}
\label{table:fast}
\end{center}
\vskip -3mm
\end{small}
\end{table}

\begin{table*}[htpb]
\vskip 3mm
\begin{center}
\begin{sc}
\resizebox{\textwidth}{!}{
\begin{tabular}{|l||ccccc|}
\hline
\diagbox{\textbf{Model}}{\hphantom{spaces}\textbf{Budget}} & $\leq$ 8KB & $\leq$ 16KB & $\leq$ 32KB & $\leq$ 64KB & $\leq$ 128KB \\
\hline
\hline
\Tstrut
Direct Conv. (3-channel) & $\mathbf{0.604}\ \textbf{[5.39KB]}$ & $\mathbf{0.629}\ \textbf{[8.65KB]}$ & $\mathbf{0.6433}\ \textbf{[19.91KB]}$ & $\mathbf{0.657}\ \textbf{[58.23KB]}$ & $\mathbf{0.657}\ \textbf{[58.23KB]}$ \Bstrut\\
\hline
\Tstrut
ProtoNN & -- & -- & $0.147$\ [24.77KB] & $0.147$\ [24.77KB] & $0.147$\ [24.77KB] \Bstrut\\
\hline
\Tstrut
Bonsai & $0.149$\ [7.88KB] & $0.153$\ [15.43KB] & $0.221$\ [30.85KB] & $0.325$\ [60.86KB] & $0.377$\ [94.52KB] \Bstrut\\
\hline
\Tstrut
FastGRNN (Row-Major) & $0.471$\ [7.57KB] & $0.515$\ [14.23KB] & $0.541$\ [31.17KB] & $0.572$\ [63.56KB] & $0.587$\ [118.50KB] \Bstrut\\
\hline
\Tstrut
FastGRNN (Channel-Major) & $0.482$\ [7.57KB] & $0.533$\ [15.80KB] & $0.553$\ [30.07KB] & $0.589$\ [63.56KB] & $0.589$\ [63.56KB] \Bstrut\\
\hline
\Tstrut
Multi-FastGRNN & $0.447$\ [7.94KB] & $0.477$\ [15.06KB] & $0.527$\ [28.75KB] & $0.558$\ [63.87KB] & $0.558$\ [124.09KB] \Bstrut\\
\hline
\end{tabular}
}
\end{sc}
\caption{Test set accuracies for methods described in \autoref{sec:background} for different memory size budgets. Actual model size given in square brackets. Bold entries denote best model for each column, i.e. memory size budget.}
\label{table:results-overview}
\end{center}
\vskip -3mm
\end{table*}

We use an Adam optimiser \cite{adam} with learning rate 0.01, which is decayed by a factor of 0.1 after 100 epochs for the 32, 64, and 128KB model candidates. Starting at the 64KB budget, we also introduce a weight decay of $5\times10^{-4}$ to reduce overfitting.
We fix the batch size to 100 and the number of epochs to 150. Due to FastGRNN's three-stage training process, we do not carry out early stopping. Instead we roll the model back to the post-compression stage epoch at which it obtained the best validation accuracy before computing the test accuracy. The best performing models after training to completion are summarised in \autoref{table:fast} and included in \autoref{table:results-overview}.




The results in \autoref{table:results-overview}
show a clear trend. Multi-FastGRNN consistently performs the worst for every size budget, and the channel-major models perform consistently the best, though only narrowly so for the 128KB budget. The fact that the channel-major models consistently beat out the row-major models suggests that FastGRNN finds inter-channel features to be more representative than intra-channel features for the CIFAR-10 data set.
Multi-FastGRNN may then be lagging behind due to the final fully-connected layer not being able to combine the intra-channel features from each unit into descriptive inter-channel features.

At the same time, a more pressing issue for Multi-FastGRNN is that compressing $U$ and $W$ is not as effective in reducing the memory footprint of the model, since each FastGRNN unit in the Multi-FastGRNN has its own prediction parameters.
In other words, for the row- and channel-major methods, the density of $U$ and $W$ provides an effective way of reducing the memory footprint of a powerful model. Meanwhile, Multi-FastGRNN has to resort to using a smaller and therefore less powerful initial model since compression is not as effective. This hypothesis is supported by the observation that the sparse models always obtained the best performance for the row- and channel-major data feed methods, regardless of size budget, while for Multi-FastGRNN the best 8KB, 16KB, and 64KB models were all dense (see \autoref{table:fast}).

Overall, \autoref{table:results-overview} shows that FastGRNN is surprisingly apt at multi-channel image classification with CIFAR-10 in memory-constrained environments, although its performance is still far from that of the Direct Convolution method.

\section{Conclusions and Future Work}
\label{sec:conclusions}

In conclusion, we have seen that the state-of-the-art methods proposed in the last few years in the resource-constrained machine learning literature vary wildly in how well they adapt to the more complex task of classifying CIFAR-10 images.
ProtoNN failed to achieve even 15\% accuracy,
despite a 76.35\% accuracy
on a 2-class version of CIFAR having been previously presented in the literature \cite{protonn}. This suggests that the ProtoNN training procedure struggles to keep up as the complexity of the task increases. The poor performance of this model may be due to the training procedure getting stuck in a local optimum.

Bonsai, which slightly outperforms ProtoNN on the 2-class version of the CIFAR data set \cite{bonsai, protonn}, did fit the data but peaked at a modest 37.7\% accuracy.
On the other end of the spectrum, FastGRNN proved surprisingly apt at multi-channel image classification, obtaining 58.9\% test set accuracy with a footprint of 63.56KB. Future work subsisting of a more thorough deep dive into FastGRNN's hyper-parameters may even push its performance in this task a bit further. At the same time, FastGRNN also proved sensitive to the way that in which input image was turned into a time series, something which we devoted significant attention to in this analysis. It appears to perform best when the data was spliced in a way which allowed it to learn inter-channel features.

Ultimately, CNNs using Direct Convolutions \cite{direct-convolutions} dominate our analysis in this paper, obtaining a 65.7\% test set accuracy with only 58.23KB of model memory usage. Future work would be to extend the analysis presented in this paper to a wider set of image classification data sets to strengthen or disprove our conclusion that CNNs dominate image classification, even in the memory-constrained domain.

Furthermore, we have focused exclusively on the memory footprints of the models, but this is not the only source of concern for embedded or constrained applications.
Future work should extend the analysis presented here with experiments investigating the latency and energy efficiency of each model, both of which are also important factors to consider in this domain.


{\small
\bibliographystyle{ieee_fullname}
\bibliography{main.bib}

\begin{thebibliography}{10}\itemsep=-1pt

\bibitem{randomsearch}
James Bergstra and Y. Bengio.
\newblock {Random Search for Hyper-Parameter Optimization}.
\newblock {\em The Journal of Machine Learning Research}, 13:281--305, 03 2012.

\bibitem{GRU}
Kyunghyun Cho, Bart van Merri{\"e}nboer, Caglar Gulcehre, Dzmitry Bahdanau,
  Fethi Bougares, Holger Schwenk, and Yoshua Bengio.
\newblock Learning phrase representations using {RNN} encoder{--}decoder for
  statistical machine translation.
\newblock In {\em Proceedings of the 2014 Conference on Empirical Methods in
  Natural Language Processing ({EMNLP})}, pages 1724--1734, Doha, Qatar, Oct.
  2014. Association for Computational Linguistics.

\bibitem{UGRNN}
Jasmine Collins, Jascha Sohl-Dickstein, and David Sussillo.
\newblock Capacity and trainability in recurrent neural networks, 2016.

\bibitem{edgemlsrc}
Don~Kurian Dennis, Yash Gaurkar, Sridhar Gopinath, Chirag Gupta, Moksh Jain,
  Ashish Kumar, Aditya Kusupati, Chris Lovett, Shishir~G Patil, and
  Harsha~Vardhan Simhadri.
\newblock {EdgeML: Machine Learning for resource-constrained edge devices}.
\newblock Retrieved January 2020.

\bibitem{protonn}
Chirag Gupta, Arun~Sai Suggala, Ankit Goyal, Harsha~Vardhan Simhadri, Bhargavi
  Paranjape, Ashish Kumar, Saurabh Goyal, Raghavendra Udupa, Manik Varma, and
  Prateek Jain.
\newblock {ProtoNN: Compressed and Accurate kNN for Resource-Scarce Devices}.
\newblock In {\em Proceedings of the 34th International Conference on Machine
  Learning-Volume 70}, pages 1331--1340, 2017.

\bibitem{cnnsrc}
Albert Gural.
\newblock {Code for reproducing work of ICML 2019 paper: Memory-Optimal Direct
  Convolutions for Maximizing Classification Accuracy in Embedded
  Applications}.
\newblock Retrieved January 2020.

\bibitem{direct-convolutions}
Albert Gural and Boris Murmann.
\newblock {Memory-Optimal Direct Convolutions for Maximizing Classification
  Accuracy in Embedded Applications}.
\newblock In Kamalika Chaudhuri and Ruslan Salakhutdinov, editors, {\em
  Proceedings of the 36th International Conference on Machine Learning},
  volume~97 of {\em Proceedings of Machine Learning Research}, pages
  2515--2524, Long Beach, California, USA, 09--15 Jun 2019. PMLR.

\bibitem{lstm}
Sepp Hochreiter and J\"{u}rgen Schmidhuber.
\newblock Long short-term memory.
\newblock {\em Neural Comput.}, 9(8):1735–1780, Nov. 1997.

\bibitem{adam}
Diederik~P. Kingma and Jimmy Ba.
\newblock {Adam: A Method for Stochastic Optimization}, 2014.
\newblock cite arxiv:1412.6980Comment: Published as a conference paper at the
  3rd International Conference for Learning Representations, San Diego, 2015.

\bibitem{cifar-10}
Alex Krizhevsky.
\newblock {Learning Multiple Layers of Features from Tiny Images}.
\newblock {\em University of Toronto}, 04 2009.

\bibitem{NIPS2012_4824}
Alex Krizhevsky, Ilya Sutskever, and Geoffrey~E Hinton.
\newblock {ImageNet Classification with Deep Convolutional Neural Networks}.
\newblock In F. Pereira, C.~J.~C. Burges, L. Bottou, and K.~Q. Weinberger,
  editors, {\em Advances in Neural Information Processing Systems 25}, pages
  1097--1105. Curran Associates, Inc., 2012.

\bibitem{bonsai}
Ashish Kumar, Saurabh Goyal, and Manik Varma.
\newblock {Resource-efficient Machine Learning in 2 {KB} {RAM} for the Internet
  of Things}.
\newblock In Doina Precup and Yee~Whye Teh, editors, {\em Proceedings of the
  34th International Conference on Machine Learning}, volume~70 of {\em
  Proceedings of Machine Learning Research}, pages 1935--1944, International
  Convention Centre, Sydney, Australia, 06--11 Aug 2017. PMLR.

\bibitem{fastgrnn}
Aditya Kusupati, Manish Singh, Kush Bhatia, Ashish Kumar, Prateek Jain, and
  Manik Varma.
\newblock {FastGRNN: A Fast, Accurate, Stable and Tiny Kilobyte Sized Gated
  Recurrent Neural Network}.
\newblock In S. Bengio, H. Wallach, H. Larochelle, K. Grauman, N. Cesa-Bianchi,
  and R. Garnett, editors, {\em Advances in Neural Information Processing
  Systems 31}, pages 9017--9028. Curran Associates, Inc., 2018.

\bibitem{mnist}
Yann Lecun, Léon Bottou, Yoshua Bengio, and Patrick Haffner.
\newblock {Gradient-based learning applied to document recognition}.
\newblock In {\em Proceedings of the IEEE}, pages 2278--2324, 1998.

\bibitem{kNN}
Antonio Mucherino, Petraq~J. Papajorgji, and Panos~M. Pardalos.
\newblock {\em {k-Nearest Neighbor Classification}}, pages 83--106.
\newblock Springer New York, New York, NY, 2009.

\bibitem{norman_2019}
Hellen Norman.
\newblock {Living on the Edge: Why On-Device ML is Here to Stay}, Apr 2019.

\bibitem{seizure-detection}
A. {Shoeb}, D. {Carlson}, E. {Panken}, and T. {Denison}.
\newblock A micropower support vector machine based seizure detection
  architecture for embedded medical devices.
\newblock In {\em 2009 Annual International Conference of the IEEE Engineering
  in Medicine and Biology Society}, pages 4202--4205, 2009.

\bibitem{simonyan2014deep}
Karen Simonyan and Andrew Zisserman.
\newblock Very deep convolutional networks for large-scale image recognition.
\newblock In {\em International Conference on Learning Representations}, 2015.

\bibitem{6322974}
J. {Viega} and H. {Thompson}.
\newblock {The State of Embedded-Device Security (Spoiler Alert: It's Bad)}.
\newblock {\em IEEE Security Privacy}, 10(5):68--70, Sep. 2012.

\bibitem{offloading-matters}
S. {Yu}, X. {Wang}, and R. {Langar}.
\newblock {Computation Offloading for Mobile Edge Computing: A Deep Learning
  Approach}.
\newblock In {\em 2017 IEEE 28th Annual International Symposium on Personal,
  Indoor, and Mobile Radio Communications (PIMRC)}, pages 1--6, 2017.

\end{thebibliography}
}

\end{document}